# Self-explainable Graph Neural Network for Alzheimer's Disease And Related Dementias Risk Prediction

Xinyue Hu, Zenan Sun, Yi Nian, Yichen Wang, Yifang Dang, Fang Li, Jingna Feng, Evan Yu, Cui Tao

# *Table of Contents*

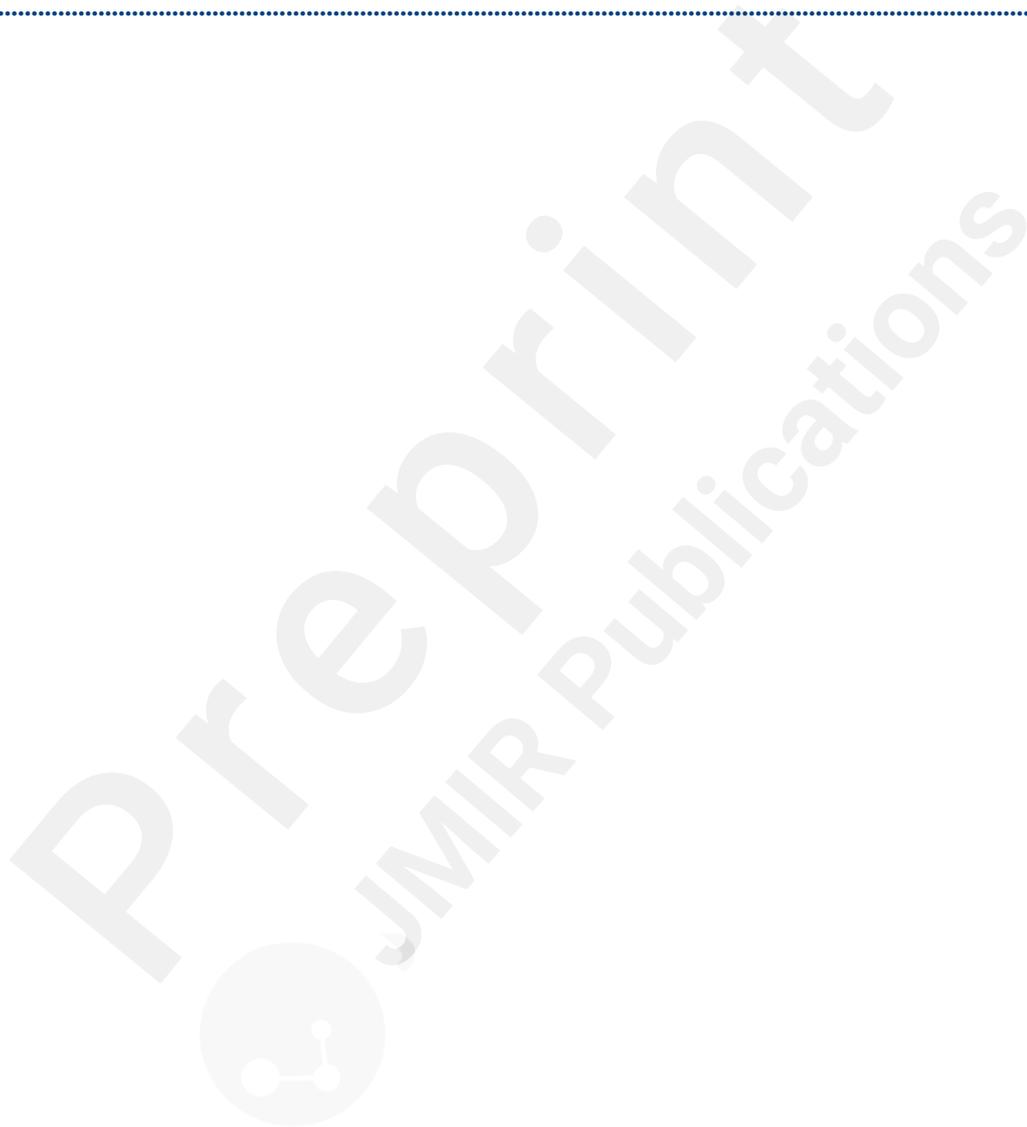

# Self-explainable Graph Neural Network for Alzheimer's Disease And Related Dementias Risk Prediction

Xinyue Hu[1, 2*]; Zenan Sun[1*]; Yi Nian[1]; Yichen Wang[3]; Yifang Dang[1]; Fang Li[1, 2]; Jingna Feng[1, 2]; Evan Yu[1]; Cui Tao[1, 2]

[1]McWilliams School of Biomedical Informatics The University of Texas Health Science Center at Houston Houston US
[2]Department of AI and Informatics Mayo Clinic Jacksonville US
[3]Division of Hospital Medicine at Perelman School of Medicine The University of Pennsylvania Philadelphia US
[*]these authors contributed equally

**Corresponding Author:**
Cui Tao
Department of AI and Informatics
Mayo Clinic
4500 San Pablo Rd S
Jacksonville
US

## *Abstract*

**Background:** Alzheimer's disease and related dementias (ADRD) ranks as the sixth leading cause of death in the US, underlining the importance of accurate ADRD risk prediction. While recent advancement in ADRD risk prediction have primarily relied on imaging analysis, yet not all patients undergo medical imaging before an ADRD diagnosis. Merging machine learning with claims data can reveal additional risk factors and uncover interconnections among diverse medical codes.

**Objective:** Our goal is to utilize Graph Neural Networks (GNNs) with claims data for ADRD risk prediction. Addressing the lack of human-interpretable reasons behind these predictions, we introduce an innovative method to evaluate relationship importance and its influence on ADRD risk prediction, ensuring comprehensive interpretation.

**Methods:** We employed Variationally Regularized Encoder-decoder Graph Neural Network (VGNN) for estimating ADRD likelihood. We created three scenarios to assess the model's efficiency, using Random Forest and Light Gradient Boost Machine as baselines. We further used our relation importance method to clarify the key relationships for ADRD risk prediction.

**Results:** VGNN surpassed other baseline models by 10% in the area under the receiver operating characteristic. The integration of the GNN model and relation importance interpretation could potentially play an essential role in providing valuable insight into factors that may contribute to or delay ADRD progression.

**Conclusions:** Employing a GNN approach with claims data enhances ADRD risk prediction and provides insights into the impact of interconnected medical code relationships. This methodology not only enables ADRD risk modeling but also shows potential for other image analysis predictions using claims data.

## Preprint Settings

1) Would you like to publish your submitted manuscript as preprint?
   ✓ **Please make my preprint PDF available to anyone at any time (recommended).**
   Please make my preprint PDF available only to logged-in users; I understand that my title and abstract will remain visible to all users.
   Only make the preprint title and abstract visible.
   No, I do not wish to publish my submitted manuscript as a preprint.
2) If accepted for publication in a JMIR journal, would you like the PDF to be visible to the public?
   ✓ **Yes, please make my accepted manuscript PDF available to anyone at any time (Recommended).**
   Yes, but please make my accepted manuscript PDF available only to logged-in users; I understand that the title and abstract will remain v

Yes, but only make the title and abstract visible (see Important note, above). I understand that if I later pay to participate in  <a href="http

**Original Manuscript**

# Self-explainable Graph Neural Network for Alzheimer's Disease And Related Dementias Risk Prediction

Xinyue Hu[1,2,3]*, Zenan Sun[2,3]*, Yi Nian[2,3], Yichen Wang[4], Yifang Dang[2,3], Fang Li[1,2,3], Jingna Feng[1,2,3], Evan Yu[2,3], Cui Tao[1,2,3]**

[1] Department of AI and Informatics, Mayo Clinic, Jacksonville, FL, USA

[2] McWilliams School of Biomedical Informatics, The University of Texas Health Science Center at Houston, Houston, TX, USA

[3] Center for Biomedical Semantics and Data Intelligence (BSDI), University of Texas Health Science Center at Houston, Houston, TX, USA

[4] Division of Hospital Medicine, Perelman School of Medicine, University of Pennsylvania, Philadelphia, PA, USA

Xinyue Hu and Zenan Sun contributed equally.

**Address for correspondence:

Cui Tao, Ph.D. Department of AI and Informatics, Mayo Clinic, Jacksonville, FL, USA; 4500 San Pablo Rd S, Jacksonville, FL 32224

E-mail: Tao.Cui@mayo.edu; Tel: 904-956-3256 Fax: 904-956-3359

## ABSTRACT

## Background

Alzheimer's disease and related dementias (ADRD) ranks as the sixth leading cause of death in the US, underlining the importance of accurate ADRD risk prediction. While recent advancement in ADRD risk prediction have primarily relied on imaging analysis, yet not all patients undergo medical imaging before an ADRD diagnosis. Merging machine learning with claims data can reveal additional risk factors and uncover interconnections among diverse medical codes.

## Objective

Our goal is to utilize Graph Neural Networks (GNNs) with claims data for ADRD risk prediction. Addressing the lack of human-interpretable reasons behind these predictions, we introduce an innovative self-explainable method to evaluate relationship importance and its influence on ADRD risk prediction.

## Methods

We employed Variationally Regularized Encoder-decoder Graph Neural Network (VGNN) integrated with our proposed relation importance method for estimating ADRD likelihood. This self-explainable method can provide a feature importance explanation in the context of ADRD risk prediction leveraging relation information within a graph. Three scenarios with 1-year, 2-year and 3-year prediction windows were created to assess the model's efficiency respectively. Random Forest and Light Gradient Boost Machine were used as baselines. By using this method, we further clarify the key relationships for ADRD risk prediction.

## Results

In scenario 1, the VGNN model showed the area under the receiver operating characteristic (AUROC) scores of 0.7272 and 0.7480 for the small subset and the matched cohort dataset. It outperforms RF and LGBM by 10.6% and 9.1% on average. In scenario 2, it achieved AUROC scores 0.7125 and 0.7281, surpassing the other models by 10.5% and 8.9%. Similarly, in scenario 3, AUROC scores of 0.7001 and 0.7187 were obtained, exceeding 10.1% and 8.5% than the baseline models. These results clearly demonstrate the significant superiority of the graph-based approach over the tree-based models (RF, LGBM) in predicting ADRD.

Furthermore, the integration of the VGNN model and our relation importance interpretation could provide valuable insight into paired factors that may contribute to or delay ADRD progression.

## Conclusion

Employing our innovative self-explainable method with claims data enhances ADRD risk prediction and provides insights into the impact of interconnected medical code relationships. This methodology not only enables ADRD risk modeling but also shows potential for other image analysis predictions using claims data.

# INTRODUCTION

## Background

Alzheimer's disease and related dementias (ADRD) currently rank as the sixth leading cause of death in the United States [1]. And currently, 47 million people live with ADRD globally [2]. By the year 2050, the prevalence of dementia is expected to triple globally [3]. These alarming statistics emphasize the pressing need of accurately predicting ADRD risk, which holds immense significance for several reasons. Firstly, it enables early detection and diagnosis, which can facilitate timely interventions and treatment plans which has the potential to slow down disease progression, improve patient outcomes, and enhance the quality of life for individuals affected by ADRD. Secondly, it also plays a crucial role in advancing research and drug development. It provides valuable insights into disease progression, risk factors, and potential therapeutic targets. By identifying individuals at high risk of developing ADRD, researchers can conduct targeted studies, clinical trials, and explore preventive measures to mitigate the impact of this debilitating disease. Thirdly, early prediction and intervention may help reduce healthcare costs associated with ADRD. By identifying individuals at risk and providing appropriate care, the burden on the healthcare system can be lessened. Nevertheless, predicting ADRD risks is an intricate task given due to its nature as a long-term chronic nature disease with multifaceted underlying causes.

In the context of ADRD risk prediction, the conventional approach predominantly involves utilizing machine learning (ML) models with medical imaging data as primary resources to achieve commendable success [4–6]. However, it is important to acknowledge that not all patients undergo routine clinical imaging tests during their regular visits, rendering medical imaging data less accessible for certain individuals. In contrast, claims data provides a more readily available data source of the ML predictors. Hence, the development of a valuable and easily trainable risk prediction tool necessitates the utilization of existing claims data as the primary input for prediction. This approach not only enhances the model's generalizability but also facilitates its adaptation to other diverse data sources.

In recent years, the emergence of graph structured data has received significant interest within the realm of deep learning [7–11]. Graphs are composed of nodes and relationships, resulting in the representation and analysis of intricate connections and patterns within the data they encapsulate. They also offer a unique combination of topological structure and individual features, which enables a rich source of information [12,13]. To analyze and model the complex relations of interconnected graph data, graph neural networks (GNNs) have emerged as a powerful tool [14]. Unlike traditional machine learning models that operate on fixed-dimensional inputs, GNNs operate directly on the graph structure which allows them to learn the representation of individuals, attributes, and relationships. In the biomedical domain, GNNs have been employed for tasks such as protein function prediction, drug discovery, disease classification, and personalized medicine [15–20]. Li et al. proposed DTI-MGNN, a framework for predicting Drug-Target Interactions (DTI) that combines a multi-channel Graph Convolutional Network and Graph Attention Network (GAT) [21]. This framework utilizes a topology graph for contextual representation, a feature graph for semantic representation, and a common representation of drug and protein pairs. DTI-MGNN has demonstrated remarkable accuracy in identifying DTIs, achieving an impressive area under the receiver operating characteristic (AUROC) score of 0.9665. Wang et al. introduced DeepDDS, a deep learning framework for predicting drug-drug interaction for the anti-cancer treatments [22]. DeepDDS uses gene expression data from the cancer cell line and the molecular graph of the drugs as input. It leverages GAT and Graph Convolution Transformer (GCT) to accurately predict the synergistic effect between drug combinations. DeepDDS has achieved an AUROC score of 0.67 on an independent test set. In the task of ADRD prediction, GCT obtained an area under the precision-recall curve (AUPRC) of 0.34 on the inpatient and outpatient EHR data from NYU Langone Health (briefly called AD-EHR) [23]. Klepl et al. integrated functional connectivity methods with GNNs to evaluate ADRD prediction performance using electroencephalography brain data. They showed that GNN-based approach outperformed CNN and SVM models and obtained an AUROC of 0.984 [24]. Zhu et al. presented VGNN, a variationally regularized encoder-decoder graph neural network, designed specifically for Electronic Health Records (EHRs) [23]. This framework showed robustness in learning graph structures by applying regularization techniques to node representations. VGNN was employed for ADRD risk prediction, and it attained an AUPRC of 0.46 when utilizing AD-EHR.

The above-mentioned GNN models [23,24] have demonstrated the potential to uncover hidden patterns, reveal biological insights, and facilitate advancements in ADRD prediction. However, the GNN architecture is a black-box model, the absence of interpretability is harmful in both users and society [25], especially in critical applications where decisions need to be explained or understood. Even though some advanced models such as GAT, GCT, and VGNN have the ability to  explain the importance of individual nodes by utilizing the attention mechanisms , they still face a limitation in their interpretability concerning the significance of underlying relationships in the prediction process. As a consequence, there is a pressing demand for research and development efforts to enhance GNNs and elucidate the influence of relationship importance in achieving more precise ADRD predictions. By addressing this interpretability issue, GNNs can become more valuable tools in advancing our understanding of Alzheimer's disease and related dementias and contributing to improved patient care and treatment strategies.

## Objective

The first focus of our study lies in the domain of risk prediction for ADRD. In this particular context, our investigation aims to utilize claims data as the sole input for our GNN-based predictive model for accurate ADRD risk prediction. We enhance the predictive power of our model by incorporating advanced graph neural network models into a framework which enables us to effectively capture the intricate relationships and dependencies inherent in the claims data.

Secondly, we introduced a novel method to assess the importance of relations within the patients' individual medical record graph and their influence on ADRD risk prediction. Generally, an additional graph explanation technique, such as GNNExplainer [26], is employed as a post-hoc method to interpret the predictions made by the graph neural networks. However, our proposed relation importance method enables an 'in-process' explanation approach which leverages the relation weights from each patient's individual graph. This method facilitates the interpretation of the graph neural network's predictions during the graph generation process itself. Besides that, our method aims to adequately calibrate the importance of each relationship within the graph, reflecting their true impact on prediction. Since, typically, when a relation connects to nodes that are highly prevalent in the graph, there is a risk of misdefining its significance. The frequent occurrence of these nodes can distort the perception of the relationship's importance, potentially leading to erroneous interpretations or biased conclusions. This bias can result in skewed importance assigned to relations and hence potentially affecting the accuracy of ADRD risk prediction. By considering the patient groups with and without ADRD, our approach helps to mitigate the potential bias resulting from node frequency, enabling a more comprehensive and reliable interpretability of relation importance for ADRD risk prediction.

## METHODS

## Cohort Description

We utilized de-identified administrative health claims data from Optum's Clinformatics® Data Mart, spanning from 2007 to 2020. This dataset comprises over 68 million patient-level enrollment records submitted by various healthcare providers, pharmacies, and other healthcare service organizations for reimbursement purposes. It is accessible for researchers through a subscription provided by the University of Texas Health Science Center (UTHealth) [27]. Approval for the use of data in this study was obtained from the UTHealth Houston Committee for the Protection of Human Subjects, under protocol HSC-SBMI-21-0965, with a waiver of consent granted.

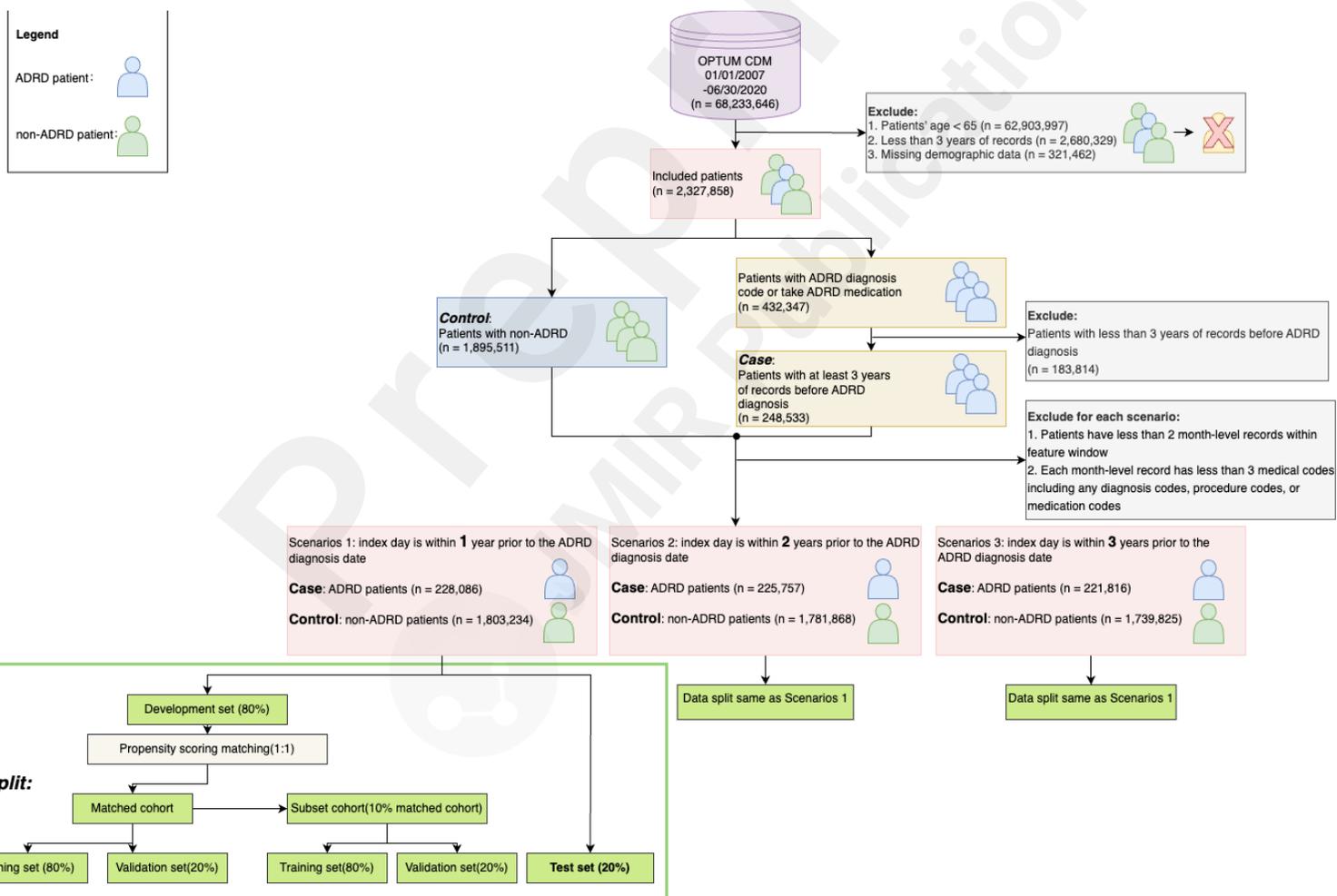

Figure 1: Overview of cohort selection for three scenarios

To construct the study cohort, several criteria were applied, as illustrated in Figure 1. Considering that ADRD primarily affects older individuals and is a chronic condition, we initially filtered out patients (n=62,903,997) who were less than 65 years old. To ensure a sufficient data history for tracking their medical conditions, patients (n=2,680,329) with a time span of less than three years between their initial and final medical records were excluded. Patients (n=321,462) who lacked demographic information were also excluded from the study. To further establish the ADRD cohort,

we employed the definition outlined in Kim's previous work [28]. Patients were classified as having

ADRD if they presented specific diagnosis codes or were prescribed relevant medications, as

detailed in Table 1. Based on the criteria mentioned above, the resulting cohort included 432,374

patients with ADRD and 1,895,511 patients without ADRD.

Table 1. ADRD diagnosis codes and medications

| ADRD Diagnosis Codes and Medication Names | |
|---|---|
| Diagnosis codes | Medication Names |
| Alzheimer's dementia – 331.0*/ G30.*<br>Vascular dementia – 290.4*/ F01.*<br>Frontotemporal dementia – 331.1*/ G31.0*<br>Lewy Body dementia – 331.82*/ G31.83<br>Senile dementia – 290.0*<br>Presenile dementia – 290.1*<br>Other specified senile psychotic – 290.8*<br>Unspecified senile psychotic condition – 290.9* | Aricept<br>Donepezil<br>Razadyne<br>Reminyl<br>Galantamine<br>Exelon<br>Rivastigmine<br>Namenda<br>Memantine<br>Acetylcholine<br>Memantine |

## Data Preprocessing

In our study, we employed a partitioning approach to categorize each patient's records into three time windows: index selection window, feature window, and prediction window (shown in Figure 2).

First, we designated a specific period before the initial diagnosis of ADRD patients or the last record for non-ADRD patients as the index selection window. In the real world, patients may seek consultations for their health conditions at any time. To simulate this visiting setting, we randomly select the index day within each patient's index selection window instead of using a fixed day. The three-year period before the index day serves as the feature window for model training purposes, while a certain period after the index day is defined as the "prediction window" for ADRD risk prediction. Additionally, we designed three scenarios with index selection windows and prediction windows of one, two, and three years in length, respectively. By employing this partitioning approach, we can comprehensively evaluate our model's predictive accuracy in dynamically predicting ADRD diagnoses. It should be noted that researchers can easily adjust the lengths of these windows to align with their specific requirements and objectives.

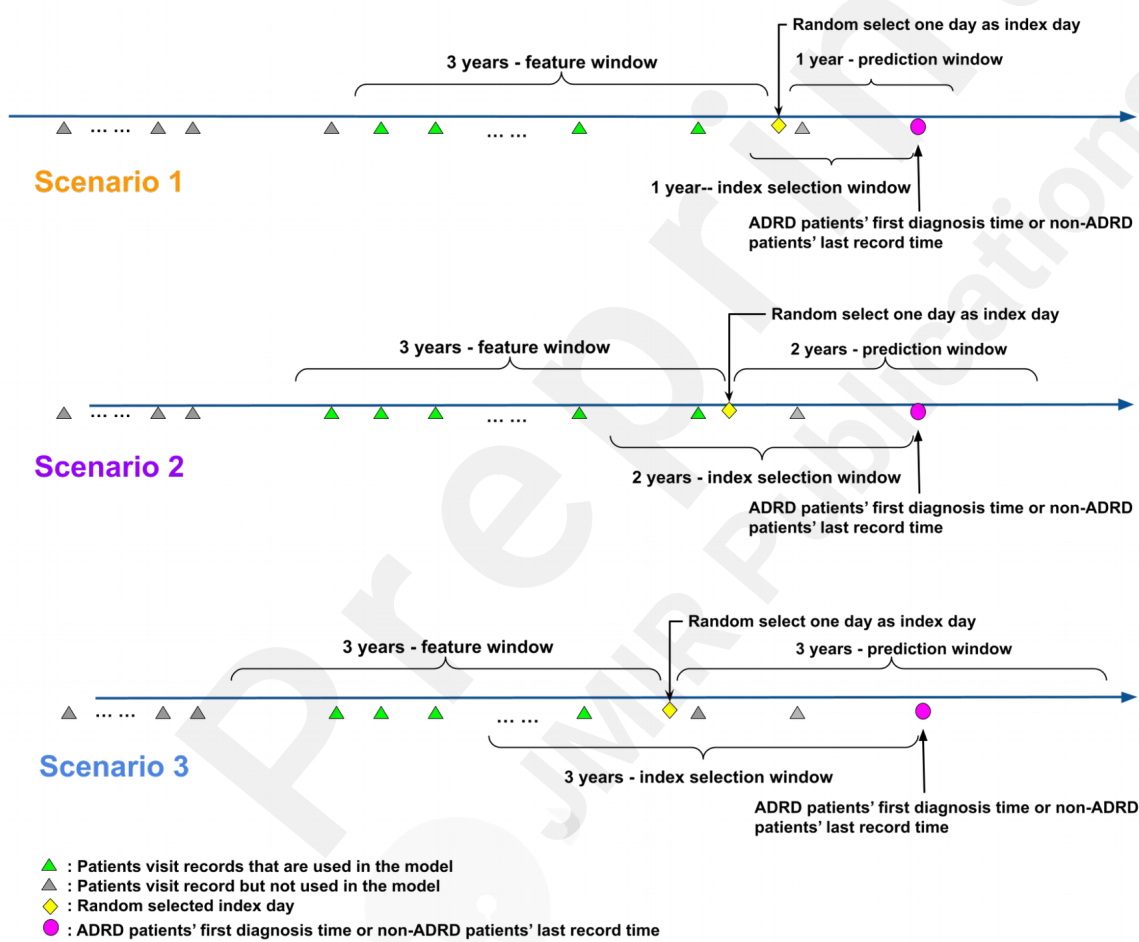

Figure 2: The definition of three scenarios

We have established a timeframe that includes an index selection window, a feature window, and a prediction window. The index selection window spanned from a specific period before the initial diagnosis of ADRD patients or the last record for non-ADRD patients. We randomly selected a day within the index selection window as the index day to simulate real-world visiting settings. The period up to three years before this index day was considered as the feature window for training the model, while the period after the index day was utilized as the prediction window. We employed 1 year, 2 years, and 3 years as the lengths of the index selection window and corresponding prediction window, respectively, to predict ADRD diagnosis dynamically.

There are also other inclusion criteria that were applied to ensure data quality and fairness of the cohort. Specifically, within the feature window, it was required that each patient had a minimum of two month-level records. Furthermore, within the records in the same month, a minimum of three medical codes (e.g., diagnosis codes, procedure codes, medication codes) needed to be present. After applying these criteria, the resulting cohort for each scenario is presented in Figure 1. In scenario 1, the cohort consisted of a total of 2,031,320 patients, comprising 228,086 ADRD patients and 1,803,234 non-ADRD patients. For scenario 2, the cohort comprised 2,007,625 patients, including 225,757 ADRD patients and 1,781,868 non-ADRD patients. Finally, in scenario 3, the cohort encompassed 1,961,641 patients, with 221,816 ADRD patients and 1,739,825 non-ADRD patients. These cohorts provide a robust foundation for further analysis and investigation in our study.

The data utilized in all cohorts included claims data consisting of diagnoses encoded with both ICD-9 and ICD-10 codes, NDC codes for pharmacy claims, and CPT and HCPCS codes for procedure. The inclusion of both ICD-10 and ICD-9 codes was necessary as the study period spanned the transition from ICD-9 to ICD-10 coding systems. All these different types of medical codes have been converted to a higher level categorization scheme to do the feature reduction and uniformity and compatibility within the study analysis. The ICD-9 and ICD-10 and the CPT and HCPCS codes are converted to Clinical Classification Software which is a tool for clustering patient diagnoses and procedures into a manageable number of clinically meaningful categories developed at the Agency for Healthcare Research and Quality (formerly known as the Agency for Health Care Policy and Research) [29]. Similarly, we are using the Pharmacologic-Therapeutic Classification System – AHFS codes to represent and group the drug NDC codes in the dataset [30]. It is a method of grouping drugs with similar pharmacologic, therapeutic, and/or chemical characteristics in a 4-tier hierarchy associated with a numeric code consisting of 2 to 8 digits. By following the conversion of these codes, the number of features was reduced from tens of thousands to hundreds. This reduction not only helps address the issue of sparsity in the model input but also improves its overall efficiency.

# Modeling

We employed the Variationally Regularized Encoder-decoder Graph Neural Network (VGNN) in combination with patients' diagnosis, procedure, and medication codes to estimate the likelihood of patients having ADRD within a designated prediction window. VGNN consists of four modules: the encoder graph, variational regularization, decoder graph, and fully connected layer. In the encoder graph module, VGNN takes three types of patients' medical codes from the feature window as input and constructs a fully connected graph comprising medical codes for each patient. The representation of each node is iteratively updated through multiple graph attention layers. To address the challenges of generating node embeddings within clusters and achieving balanced attention weights, VGNN incorporates a variational regularization layer. This layer helps prevent model collapse and maintains the model's expressive capacity. The decoder graph module utilizes the node representations generated by the encoder graph and the variational regularization layer to compute the weighted relations between each node. These weighted relations effectively capture the relationships among different medical codes. And finally, a linear feed-forward layer is used to calculate the probability and produce the binary classification for identifying an individual having ADRD.

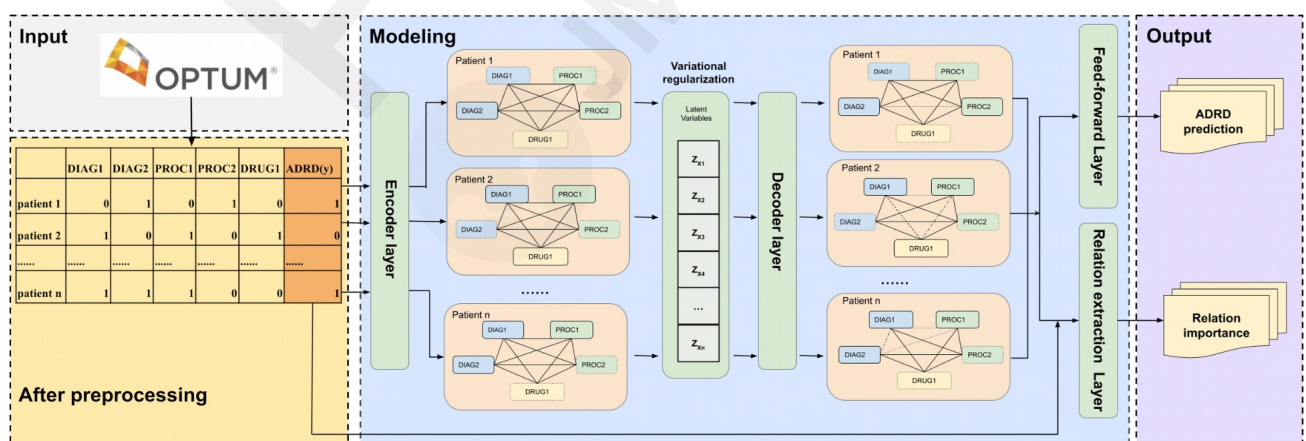

Figure 3: The workflow of our study pipeline including data preprocessing, graph modeling, and final output

We employed VGNN to predict the likelihood of ADRD using patients' medical records sourced from Optum Clinformatics®. The data was input into the encoder layer of VGNN, generating a fully connected graph specific to each patient. The variational regularization layer was then applied to prevent issues like mode collapse and maintain the model's capacity to represent information effectively. Additionally, the decoder graph module used node representations to compute weighted relations between nodes, which effectively captured relationships among different medical codes. Finally, a linear feed-forward layer was utilized to calculate probabilities and perform binary classification.

We initiated the modeling process by reserving 20% of patients from the entire dataset for testing purposes. Given that ADRD is more prevalent in the elderly population [1] and our dataset exhibits high imbalance, we employed the propensity score matching method based on age and gender to mitigate potential biases associated with these factors. This matching process ensured that our model's input cohort consisted of individuals with similar age and gender distributions, reducing potential confounding effects associated with these variables. As a result, we created a balanced cohort with a one-to-one ratio of control and case groups from the remaining 80% of the entire dataset. This downsampling approach is a popular method in clinical research to create a balanced covariate distribution between treated and untreated groups which could help significantly improve the model's ability to handle imbalanced data [31]. We named it as the matched cohort and employed it for the purposes of model development and validation. Additionally, we generated a smaller subset named as subset cohort which is 10% of the matched cohort. This action allows us to evaluate the model's performance on a smaller-scale dataset effectively. In order to assess the efficacy of our approach, we built models for three different scenarios. Moreover, we employed Random Forest (RF) and Light Gradient Boost Machine (LGBM) as baseline models and compared their performance with that of VGNN. The overall workflow of our model pipeline is shown in Figure 3.

## Relation Importance

After the completion of model training, we then used the trained model to build the interconnected medical record graph for each individual patient. In order to evaluate the significance of various relationships in ADRD prediction, we extracted adjacency matrices $A = \{A_i : i \in 1, \ldots, N\}$ from the medical graphs of N patients in the training set of the matched cohort. The values within these adjacency matrices serve as indicators of the relation importance associated with predicting ADRD. Given that the generated graphs are directional, the adjacency matrices $A$ are not symmetric. Therefore, we took an additional step to mitigate the influence of directionality by computing the average of the original adjacency matrix and its transposed matrix. Then, the updated adjacency matrix is:

$$A_i = \frac{1}{2}\left(A_i + A_i{}'\right), \forall\, i \in 1,\ldots,N \ . \tag{1}$$

This adjacency matrix enables us to gain insights into the intricate relationships between medical codes and their predictive power for ADRD.

However, it is crucial to consider that medical codes with higher frequencies may have received relatively larger weights compared to others, potentially introducing bias in the analysis. Given that $A^{+¿}$ are the adjacency matrices of ADRD patient case group and $A^{-¿}$ are the adjacency matrices of ADRD patient control group, we calculated the mean adjacency matrix of these two patient groups as:

$$A^{+¿} = \bar{MEAN}¿¿ \tag{2}$$

$$A^{-¿} = \bar{MEAN}¿¿ \tag{3}$$

By subtracting the negative mean adjacency matrix from the positive mean adjacency matrix, we eventually obtained a mean weight-difference matrix:

$$W = A^{+¿} - \bar{A}^{-¿}¿¿ \tag{4}$$

This mean weight-difference matrix $W$ captured the relative significance of the medical code weights. A higher positive value inside $W$ indicates a greater importance in predicting ADRD, while a lower negative value suggests a reduced likelihood of ADRD occurrence. A value of 0 in $W$ means that the relation does not affect a patient getting ADRD.

# RESULTS

## Hyperparameter Setting

We trained the VGNN model with the following hyperparameters: learning rate of 0.0001, batch size of 128, and a dropout rate of 0.1. We utilized Adam optimizer for gradient descent and trained the model for 200 epochs. The model consisted of 2 graph layers and 1 attention head. To balance the binary cross-entropy and Kullback-Leibler divergence losses, a parameter value of 0.002 was utilized. Additionally, edge information was extracted after the attention layer to facilitate future calculations of relation importance. Additionally, we used the grid search method to tune the RF and LGBM baseline models. The hyperparameters for RF and LGBM are n_estimators=100, min_samples_split=2, min_samples_leaf=1, and n_estimators=300, boosting_type='gbdt', num_leaves=31, learning_rate=0.1 respectively.

## Performance Evaluation

To evaluate the performance of each model, we employed AUROC as a measurement. As shown in Figure 4, the VGNN model achieved AUROC scores of 0.7272 and 0.7480 for the subset cohortand the matched cohort, respectively in scenario 1. It outperformed the RF and LGBM models by an average of 10.6% and 9.1% across the two datasets. For scenario 2, the VGNN model obtained AUROC scores of 0.7125 and 0.7281 for the subset cohort and matched cohort, respectively. It exhibited superior performance compared to the other two models by an average of 10.5% and 8.9% across the two datasets. Finally, in scenario 3, the VGNN model achieved AUROC scores of 0.7001 and 0.7187, which surpassed by an average of 10.1% and 8.5% across the two datasets. The results clearly demonstrate that the GNN approach (VGNN) outperforms the tree-based models (RF, LGBM) significantly in predicting ADRD.

Furthermore, we identified the five most important relations for both positive and negative predictions of ADRD in Table 2. Among the top 5 negative highest-weighted relationships, "Neoplasms Of Unspecified Nature Or Uncertain Behavior" exhibits its influence across all relations within Scenario 1, "Consultation, Evaluation, And Preventative Care" makes a total of four appearances within Scenario 2, while "Quinolone Antibiotics" spans all relations in Scenario 3. Within the set of top 5 positive highest-weighted relationships, both "Routine Chest X-Ray" and "Electrocardiogram" appear three times each in Scenario 1, "Substance-Related Disorders" contributes to four relationships in scenario 2, and "Substance-Related Disorders" emerges as the most frequently medical code in Scenario 3.

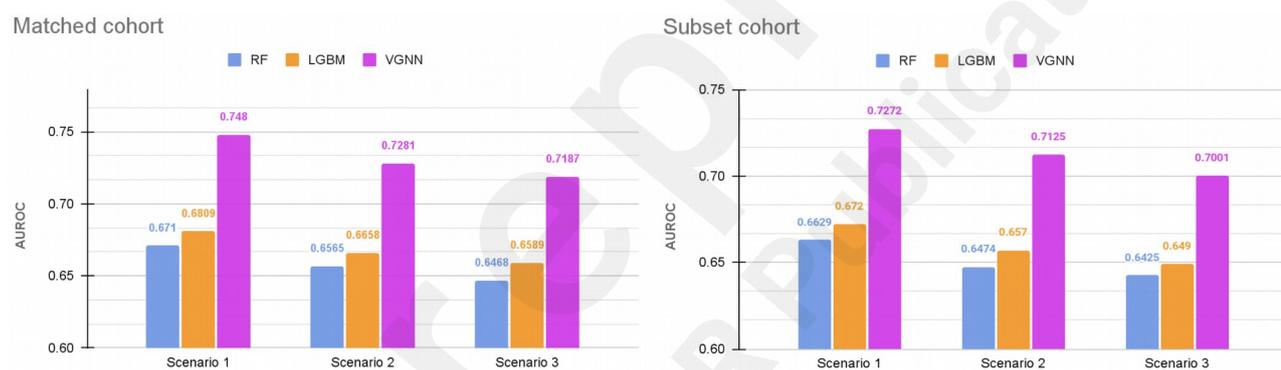

Figure 4: The model performance (AUROC scores) for ADRD risk prediction

Matched cohort: employ 1:1 propensity score match for case and control in the original training data (i.e., 80% of full data), to achieve a balanced dataset and train the models, and test the models in the hold-out 20% of full data
Subset cohort: use around 10% of the Matched cohort (i.e., 20,000 for both case and control) to train the models, and test the models in the hold-out 20% of full data.

Table 2. Top 5 positive highest-weighted relations and top 5 negative highest-weighted relations

| | | Scenario 1 | | Scenario 2 | | Scenario 3 | |
|---|---|---|---|---|---|---|---|
| Top 5 negative highest-weighted relations | Neoplasms Of Unspecified Nature Or Uncertain Behavior | Consultation, Evaluation, And Preventative Care | Consultation, Evaluation, And Preventative Care | Dihydropyridines | Quinolone Antibiotics | Suture Of S And Subcutane Tissue |
| | Neoplasms Of Unspecified Nature Or Uncertain Behavior | Lens And Cataract Procedures | Consultation, Evaluation, And Preventative Care | Diseases Of White Blood Cells | Quinolone Antibiotics | Lens And Cata Procedures |
| | Neoplasms Of Unspecified Nature Or Uncertain Behavior | Hyperlipidemia | Consultation, Evaluation, And Preventative Care | Upper Gastrointestinal Endoscopy, Biopsy | Quinolone Antibiotics | Essential Hypertension |
| | Neoplasms Of Unspecified Nature With Complications | Diabetes Mellitus | Consultation, Evaluation, And Preventative Care | Other Ct Scan | Quinolone Antibiotics | Diagnostic Ultrasound Head And Neck |
| | Neoplasms Of Unspecified Nature Or Uncertain Behavior | Diagnostic Ultrasound Of Head And Neck | Diseases Of White Blood Cells | Dihydropyridines | Quinolone Antibiotics | Psychological A Psychiatric Evaluation A Therapy |
| Top 5 positive highest-weighted relations | Routine Chest X-Ray | Electrocardiogram | Substance-Related Disorders | Electrocardiogram | Schizophrenia And Other Psychotic Disorder | Substance-Relat Disorders |
| | Routine Chest X-Ray | Other Laboratory | Substance-Related Disorders | Other Laboratory | Schizophrenia And Other Psychotic Disorder | Diagnostic Procedures Nose, Mouth A Pharynx |
| | Routine Chest X-Ray | Heart Valve Disorders | Substance-Related Disorders | Routine Chest X-Ray | Diagnostic Ultrasound Of Head And Neck | Arthrocentesis |
| | Electrocardiogram | Other Laboratory | Electrocardiogram | Inguinal And Femoral Hernia Repair | Diagnostic Ultrasound Of Head And Neck | Substance-Relat Disorders |
| | Electrocardiogram | Heart Valve Disorders | Substance-Related Disorders | Coronary Atherosclerosis And Other Heart Disease | Substance-Related Disorders | Other Diagno Radiology A Related Techniq |

## DISCUSSION

## Principal Findings

Based on our study results, we found that some potential candidates might be relevant to AD risk prediction and treatment. Our self-explainable GNN prediction method reveals the underneath connections between medical codes for ADRD risk prediction. Some code pairs have shown to accelerate ADRD progression, while others exhibit potential to slow down its development. When implementing our relation importance interpretation method, the GNN results are explainable, setting it apart from other deep learning models. Moreover, several code pairs extracted from the GNN align with findings from previous research. Those code pairs which are not proved could offer valuable insights beyond the scope of current studies, opening up avenues for further investigation and enhancing our understanding of ADRD risk prediction. Table 2 shows top 5 positive highest-weighted relations and the top 5 negative highest-weighted relations. In the following sections, we will present examples of code pairs derived from the GNN model results, and highlight their significance based on validated evidence from prior studies.

Our study found that certain pairs of medical codes can be associated with a decreased likelihood of ADRD diagnosis. For instance, the treatment of more acute conditions, such as cancer or neoplasms, may delay the diagnosis of ADRD. We hypothesize that "Neoplasms Of Unspecified Nature Or Uncertain Behavior" may be correlated with higher healthcare utilization or more frequent physician visits, similar to the code "Consultation, Evaluation, and Preventative Care". The co-occurrence of these two types of coding could potentially lower the risk of ADRD. Regular healthcare visits could potentially reduce the risk of ADRD by improving modifiable risk factors and mitigating social isolation in elderly patients. Lee's work in 2021 revealed that cataract extraction is linked to a reduced risk of developing dementia among older adults [32]. Cataract extraction has been associated with enhanced engagement in intellectually stimulating activities, such as reading and video consumption, as well as increased physical activity. These changes in lifestyle and cognitive engagement following cataract surgery may contribute to a delay in the onset of ADRD. Consequently, the second node pair involving "Neoplasms Of Unspecified Nature Or Uncertain Behavior" and "Lens And Cataract Procedures" also holds relevance and supports the observed association. In scenario 2, Peters et al. have indicated that the use of calcium channel blockers, specifically dihydropyridines, is associated with a lower decline in cognitive function compared to other hypertensive treatments [33]. Thus, the presence of the "Consultation, Evaluation, And Preventative Care" and "Dihydropyridines" nodes pair ranking first in importance is consistent with the reported associations. The most frequently appeared node in scenario 3 is "Quinolone Antibiotics". According to the study by Pham TDM and his colleagues, it is a class of medication commonly prescribed to treat various bacterial infections and is primarily used for their antimicrobial properties [34]. And from Gao's work, their review study indicates that the brain inflammation caused by microbial infections may be one of the etiologies of ADRD, and antibiotics as novel treatments may be beneficial for delaying the development of ADRD [35]. Quinolones exhibit a distinct pharmacokinetic profile characterized by a higher cerebrospinal fluid to serum concentration ratio compared to other commonly prescribed antibiotics [36]. This unique attribute may underlie the observed robust negative correlation between quinolone administration and the development of ADRD, distinguishing its potential protective effect from that of other antibiotics. The utilization of quinolones likely correlates with younger age, as its use in elderly adults is less often due to the increased risk of tendon rupture. However, this is less likely to explain its negative correlation with the onset of ADRD in our age-matched cohorts. So, in other words, it can be hypothesized that "Quinolone Antibiotics" may potentially exhibit a slowing effect on the progression of ADRD. Combined with the aforementioned node "Lens And Cataract Procedures", the observed association of this node pair holds validity and worth further investigation.

Our study also found certain medical codes to be positively associated with a higher likelihood of ADRD diagnosis. This can be explained by the fact that Alzheimer's disease, to a certain degree, is a "diagnosis of exclusion". Procedures like "Routine Chest X-Ray" and "Electrocardiogram" are commonly used as initial steps in diagnosing altered mental status, which is often the first sign of ADRD. Chest X-Ray is often used to rule out any underlying pneumonia, while electrocardiogram may be used to rule out arrhythmia [37]. Similarly, "Diagnostic Ultrasound Of Head And Neck" is commonly done to rule out conditions like carotid artery clot, stenosis, or plaque in the setting of stroke workups. Once patients begin to verify these initial diagnoses of altered mental status, they are more likely to undergo comprehensive and relevant testing to exclude other potential causes of the symptoms, which may potentially lead to a timely determination of ADRD. Several researches have also found that alcohol and drug use could affect mental state and cognitive function [38]. People who abuse intoxicating substances for a considerable period may bring on dementia or accelerate the neurological damage associated with Alzheimer's [39].

From the modeling aspect, to the best of our knowledge, our approach offers distinct advantages in comparison to previous studies on early diagnosis of ADRD with or without GNN methods. For instance, Li utilized a gradient boost tree and logistic regression to assess ADRD risk using EHR data from the OneFlorida+ Clinical Research Consortium [40]. They identified significant clinical and social factors through SHAP values; however, these factors were commonly known risk factors. In contrast, our findings unveil potential risk factors and explain the interaction among these factors in ADRD prediction. While VGNN demonstrates good interpretability by showcasing attention weights among features, it fails to explain how these features positively or negatively impact ADRD prediction [23]. On the other hand, our model offers interpretations of potential risk factors and illustrates their influence on outcomes. Furthermore, our proposed self-explainable framework mitigates the potential bias resulting from the prevalence of medical codes. Klepl et al. conducted EEG-based ADRD prediction using GNN methods [24]. As medical image data is unavailable for every patient during routine examinations, the limitations arise due to the restricted user cases. Furthermore, they only assessed model performance against baseline models without providing any feature interpretation. Conversely, our method enhances interpretability by leveraging more accessible data, thereby promoting broader applicability and understanding in ADRD prediction. Overall, this is the first work that proposed a self-explainable framework, providing a feature importance explanation in the context of ADRD risk prediction leveraging relation information within a graph. Compared with other studies in ADRD risk predictions, our method can directly interpret the relation importance within the training process. It does not require any additional post-hoc explanation method such as GNNExplainer [26]. In other words, within our framework, it takes no additional time to get an explanation.

In summary, we showed that using the GNN approach for ADRD prediction has better performance compared to baseline models. Moreover, with the incorporation of our relation importance method, the model's results become explainable, providing valuable insights into the underlying factors contributing to ADRD risk prediction.

## Limitations

Our prediction does not incorporate time information into the modeling process. In our current study, we aggregated three years of records into a single representation and treated them equally without considering their temporal sequence. In the real-world clinical setting, medical events, procedures, or medications obtained at different times should carry different levels of significance. In other words, events occurring closer in time to the prediction window are expected to have a greater impact on the disease prediction. In our future study, we could employ a time series model and/or positional encoding method to establish connections between patients' multiple visit records for more accurate predictions and provide more valuable insights of ADRD prediction.

On the other hand, it's important to note that certain predicted correlations may not causally assist clinicians in diagnosing ADRD. For instance, initiating tests for early detection of altered mental status might lead patients to identify ADRD through various related tests. Nonetheless, from the clinician's aspect, ordering these test results may not be helpful for early ADRD prediction. In our future work, we could try to exclude these "subjective patient-related factors" and instead focus on investigating more objective risk factors that could potentially influence the prediction of ADRD.

## Conclusion

In this article, we employed an advanced self-explainable GNN approach and developed a relation importance interpretation method for ADRD risk prediction task based on claims data source. The VGNN model's effectiveness was evaluated across three distinct scenarios, with comparisons made against RF and LGBM machine learning models. The model performance achieved satisfactory results. In addition, we provided the interpretation for the node pairs extracted from the KG which was generated from the VGNN model. Furthermore, we demonstrated the results' future applicability and explained the important node pairs which align with prior research findings. This work contributes to the advancement of ADRD prediction models and reinforces the importance of interpretable results for informed clinical decision-making and early detection, etc.

## Abbreviation

ADRD: Alzheimer's disease and related dementia
AUPRC: Area under the precision-recall curve
AUROC: Area under the receiver operating characteristic curve
DTI: Drug-target Interaction
EHR: Electronic health record
GAT: Graph attention network
GCT: Graph convolution transformer
GNN: Graph neural network
LGBM: Light gradient boost machine
ML: Machine learning
RF: Random forest
VGNN: Variational graph neural network

## Acknowledgements

We thank Luyao Chen and Xiaoqian Jiang for extracting data from Optum's Clinformatics®.

## Author Contributions

C.T. conceived the research project. X.H, Z.S., and Y.N. designed the pipeline and method. X.H., Z.S., and Y.N. implemented the deep learning model of the study and the explanation method. X.H., Z.S. and Y.W. prepared the manuscript. X.H. and Y.D. provided the cohort selection, scenarios definition, model workflow and model performance figures. X.H. and J.F. prepared the data. F.L. and E.Y. provided suggestions on data filtering, index day selection and model design. Y.W. composed clinical discussion on the principle finding. C.T. supervised the research and critically revised the manuscript. All authors proofread the paper and provided valuable suggestions. All the authors have read and approved the final manuscript.

## Funding Statement

This research was partially supported by NIH grants under Award Numbers R01AG084236, R01AG083039, RF1AG072799 and R56AG074604.

## Conflict of Interest

The authors declare no competing interests.

using real-world electronic health records. *Alzheimers Dement*. 2023;19:3506–18

# Supplementary Files

# Figures

Overview of cohort selection for three scenarios.

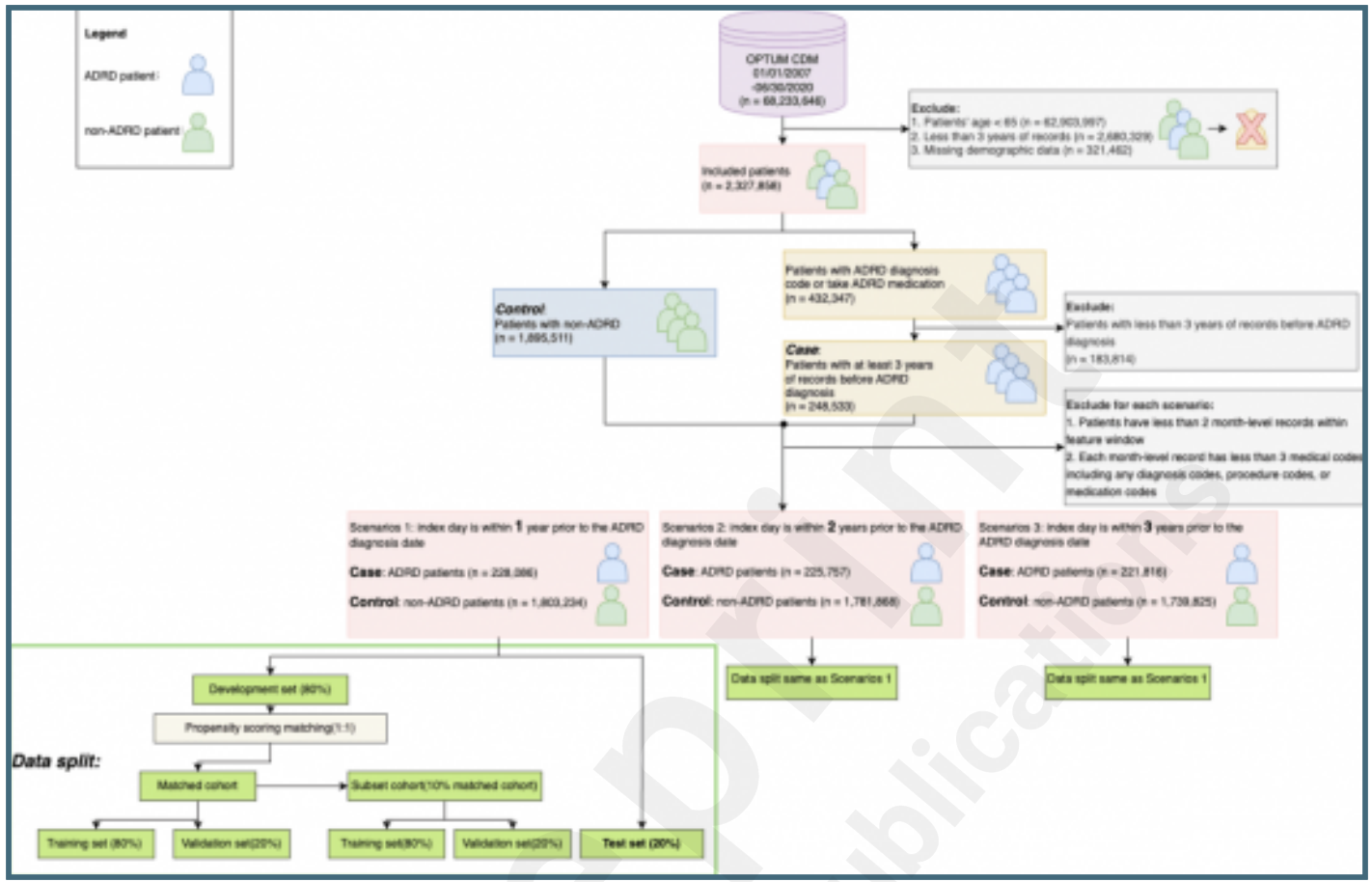

The definition of three scenarios.

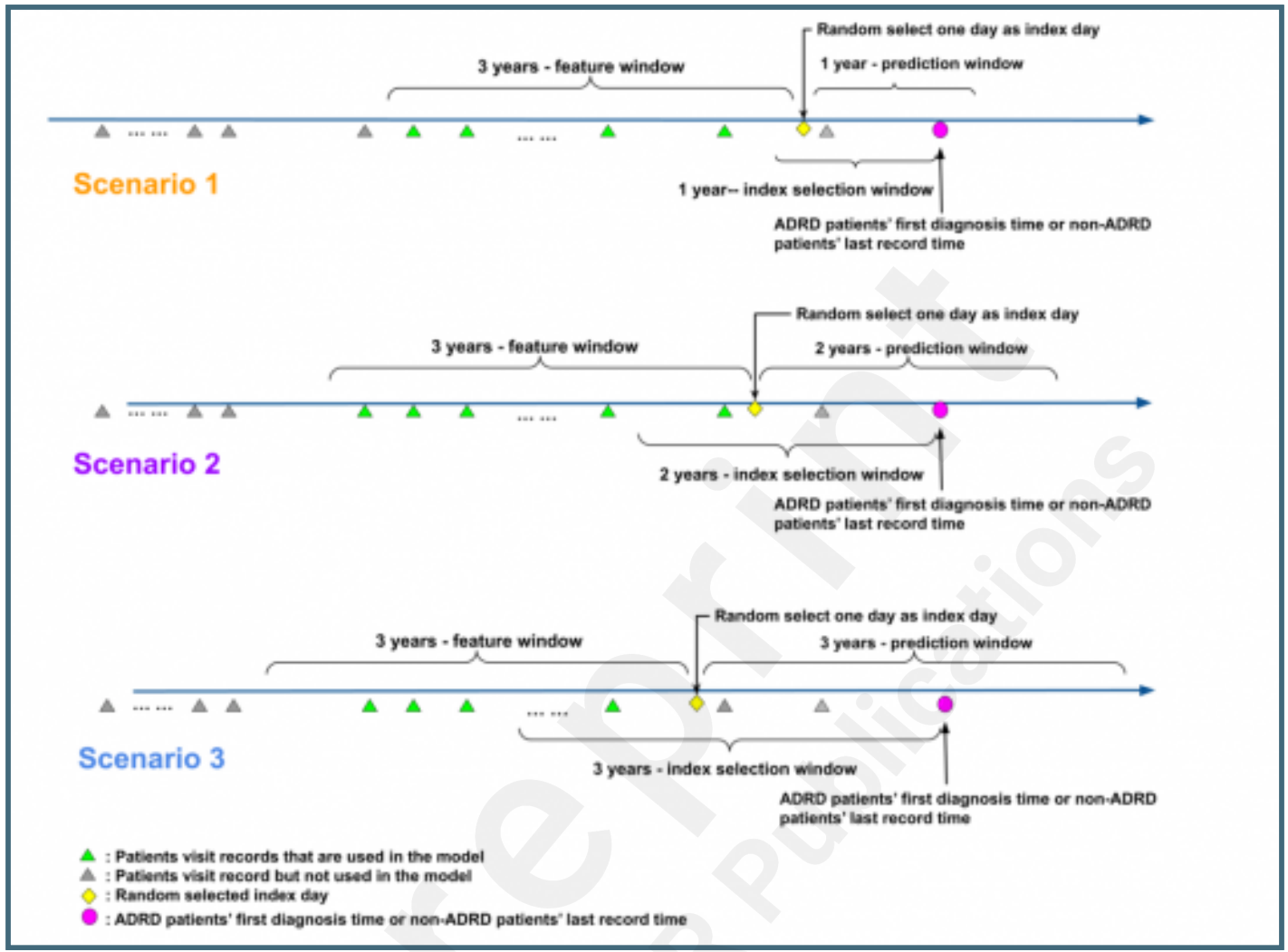

The workflow of our study pipeline including data preprocessing, graph modeling, and final output.

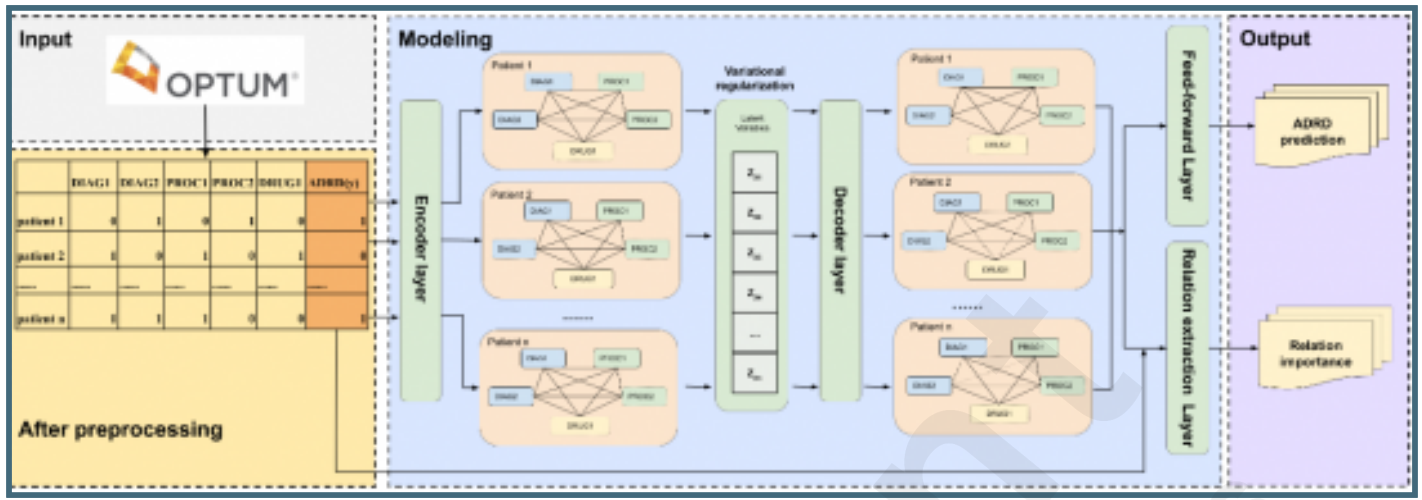

The model performance (AUROC scores) for ADRD risk prediction.

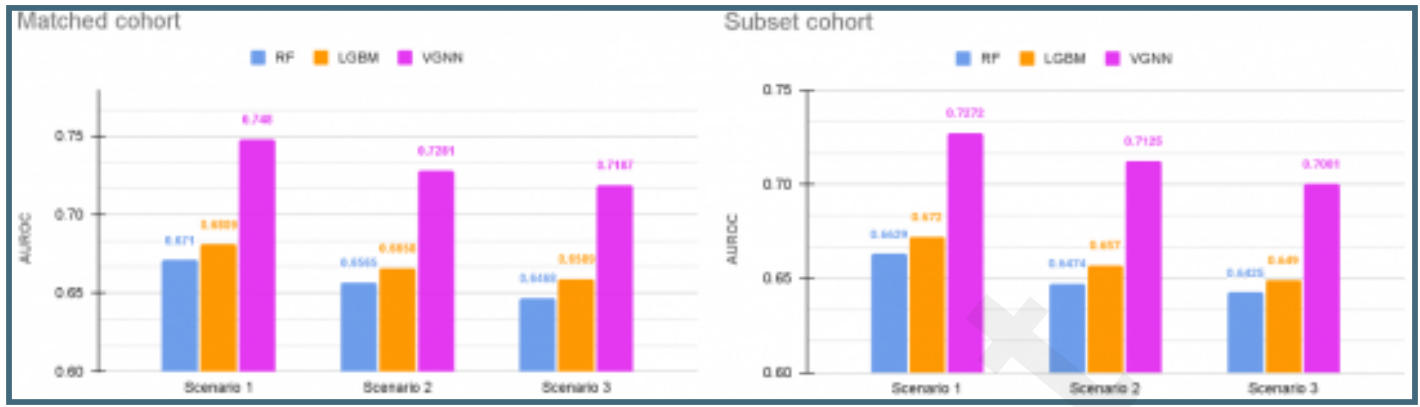